\documentclass[11pt]{article}
\usepackage{coling2016}
\usepackage{times}
\usepackage{url}
\usepackage{latexsym}

\usepackage{CJK}
\usepackage{amsfonts}
\usepackage{amsmath}
\usepackage{amssymb}
\usepackage{amsthm}
\usepackage{multirow}
\usepackage{algorithm}
\usepackage{algorithmic}
\usepackage{subfigure}
\usepackage[dvips]{graphicx}
\usepackage[usenames]{color}
\usepackage{tikz-dependency}

\makeatletter
\newcommand\argmax{\mathop{\operator@font arg~max}}
\newcommand\argmin{\mathop{\operator@font arg~min}}
\makeatother

\graphicspath{{../fig/}}

\title{Training Dependency Parsers with Partial Annotation}

\author{Zhenghua Li, ~~ Yue Zhang,~~ Jiayuan Chao, ~~ Min Zhang\thanks{~~Correspondence author} \\
Soochow University, Suzhou, China \\
{\tt \{zhli13,minzhang\}@suda.edu.cn, zhangyue1107@qq.com}
}

\date{}

\begin{document}

\maketitle

\begin{abstract}

Recently, these has been a surge on studying how to obtain partially annotated data for model supervision.
However, there still lacks a systematic study on how to train statistical models with partial annotation (PA).
Taking dependency parsing as our case study, this paper describes and compares two straightforward approaches for three mainstream dependency parsers.
The first approach is previously proposed to directly train a log-linear graph-based parser (LLGPar) with PA based on a forest-based objective.
This work for the first time proposes the second approach to directly training 
a linear graph-based parse (LGPar) and a linear transition-based parser (LTPar)
with PA based on the idea of constrained decoding.
We conduct extensive experiments on Penn Treebank under three different settings for simulating PA, i.e., 
random dependencies, most uncertain dependencies, and dependencies with divergent outputs from the three parsers.
The results show that LLGPar is most effective in learning from PA and LTPar lags behind the graph-based counterparts by large margin.
Moreover, LGPar and LTPar can achieve best performance by using LLGPar to complete PA into full annotation (FA).


\end{abstract}

\section{Introduction}\label{sec:intro}

\begin{figure}[b]
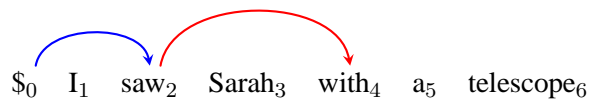

\begin{center}
\begin{dependency}[arc edge, arc angle=80, text only label, label style={above}] 
\begin{deptext} [row sep=0.4cm, column sep=.22cm] 
\$$_0$ \& I$_1$ \& saw$_2$ \& Sarah$_3$ \& with$_4$ \& a$_5$ \& telescope$_6$ \\
\end{deptext}
\depedge[edge style={blue, thick}]{1}{3}{}
\depedge[edge style={red, thick}]{3}{5}{}
\end{dependency}
\caption{
An example partial tree, where only the heads of ``$\textsl{saw}$'' and ``$\textsl{with}$'' are given.
}
\label{fig:example-partial-tree} %
\end{center}
\end{figure}

Traditional supervised approaches for structural classification assume full annotation (FA), meaning that the training instances have complete manually-labeled structures. 
In the case of dependency parsing, FA means a complete parse tree is provided for each training sentence.
However, recent studies suggest that it is more economic and effective to construct labeled data with partial annotation (PA). A lot of research effort has been attracted to obtain partially-labeled 
data for different tasks via active learning \cite{sassano-p10-partial-annotation,mirroshandel-iwpt-2011-partial-annotation,shoushan-c12-active-learning-word-boundary,marcheggiani-d14-active-learning-partial-annotation,flannery-iwpt15-active-learning-partial-annotation,zhenghua-p16}, cross-lingual syntax projection \cite{spreyer-kuhn:2009:CoNLL,ganchev-gillenwater-taskar:2009:ACLIJCNLP,jiang-wenbin-p10-bilingual-projection,zhenghua-c14}, or mining natural annotation implicitly encoded in web pages \cite{jiang-p13-natural-annotation,liu-d14-crf-natural-annotation,nivre-j14-constrained,yang-vozila-d14-crf-partial-annotation}.  
Figure \ref{fig:example-partial-tree} gives an example sentence partially annotated with two dependencies.

However, there still lacks systematic study on 
how to train structural models such as dependency parsers with PA.
Most previous works listed above rely on ad-hoc strategies designed for only basic dependency parsers.
One exception is that \newcite{zhenghua-c14} convert partial trees into forests and train a log-linear graph-based dependency parser (LLGPar) with PA based on a forest-base objective, showing promising results. 
Meanwhile, it is still unclear how PAs can be used to train state-of-the-art linear graph-based (LGPar) and transition-based parser (LTPar).
Please refer to Section \ref{sec:related-work} for detailed discussions of previous methods for training parsers with PA.

This paper aims to thoroughly study this issue and make systematic comparison on different approaches of training parsers with PA. 
In summary, we make the following contributions.
\begin{itemize}
\item We present a general framework for directly training state-of-the-art LGPar and LTPar with PA based on \emph{constrained decoding}.
The basic idea is to use the current feature weights to parse the sentence under the PA-constrained search space, and use the best parse as a pseudo gold-standard reference for feature weight update during perceptron training. We also implement the forest-objective based approach of \newcite{zhenghua-c14} for LLGPar.
\item We have made thorough comparison among different \emph{directly-train} approaches under three different settings for simulating PA, i.e., random dependencies, most uncertain dependencies, and dependencies with divergent outputs from the three parsers. We have also compared the proposed directly-train approaches with the straightforward \emph{complete-then-train} approach. 
\item Extensive experiments on Penn Treebank lead to several interesting and clear findings. 
\end{itemize}



\section{Dependency Parsing}\label{sec:dp}

Given an input sentence $\mathbf{x}=w_0w_1...w_n$, dependency parsing builds a 
complete dependency tree $\mathbf{d}$ rooted at $w_0$, where 
$w_0$ is an artificial token linking to the root of the sentence \cite{book-dp-09}.
A dependency tree comprises a set of dependencies, namely 
$\mathbf{d}=\{h \curvearrowright m: 0 \leq h \leq n, 1 \leq m \leq n\}$, 
where $h \curvearrowright m$ is a dependency from a \emph{head} word $h$ to a \emph{modifier} word $m$. 
A complete dependency tree contains $n$ dependencies, namely $|\mathbf{d}|=n$, whereas
a partial dependency tree contains less than $n$ dependencies, namely $|\mathbf{d}|<n$.
Alternatively, FA can be understood as a special form of PA.
For clarity, we denote a complete tree as $\mathbf{d}$ and  a partial tree as $\mathbf{d}^p$.
The decoding procedure aims to find an optimal complete tree $\mathbf{d}^*$:
\begin{equation}\label{eq:optimal-tree}
\begin{split}
& \mathbf{d}^* = \argmax_{\mathbf{d} \in \mathcal{Y}(\mathbf{x})} {Score(\mathbf{x}, \mathbf{d}; \mathbf{w})} \\
& Score(\mathbf{x},\mathbf{d}; \mathbf{w}) = \mathbf{w} \cdot \mathbf{f}(\mathbf{x},\mathbf{d})
\end{split}
\end{equation}
where $\mathcal{Y}(\mathbf{x})$ defines the search space containing all legal trees for $\mathbf{x}$;
$Score(\mathbf{x}, \mathbf{d}; \mathbf{w})$ is a score/probability of $\mathbf{d}$;
$\mathbf{f}(\mathbf{x},\mathbf{d})$ is a sparse accumulated feature vector corresponding to $\mathbf{d}$; $\mathbf{w}$ is the feature weight vector.



\subsection{Graph-based Approach}\label{sec:graph-dp}

To facilitate efficient search, the graph-based method factorizes the score of a dependency into those of small subtrees $\mathbf{p}$: 
\begin{equation}
Score(\mathbf{x},\mathbf{d}; \mathbf{w})
= \sum_{\mathbf{p} \subseteq \mathbf{d}}{Score(\mathbf{x}, \mathbf{p}; \mathbf{w})}
\end{equation}
Dynamic programming based \emph{exact search} are usually applied to find the optimal tree \cite{mcdonald-acl05-proj,McDonald-eacl06-non-proj-2-order,carreras:2007:EMNLP-CoNLL2007,koo-acl10-3o}.
We adopt the second-order model of \newcite{McDonald-eacl06-non-proj-2-order} which incorporates two kinds of subtrees, i.e., single dependencies and adjacent siblings, and the feature set described in \newcite{bohnet-C10-hash-kernel}.

\textbf{A log-linear graph-based parser (LLGPar)} defines
the conditional probability of $\mathbf{d}$ given $\mathbf{x}$ as
\begin{equation}\label{eq:prob-tree}
p(\mathbf{d}|\mathbf{x};\mathbf{w})  = \frac{e^{Score(\mathbf{x},\mathbf{d};\mathbf{w})}}{\sum_{\mathbf{d'} \in \mathcal{Y}(\mathbf{x})} {e^{Score(\mathbf{x},\mathbf{d'};\mathbf{w})}}} 
\end{equation}
For training, $\mathbf{w}$ is optimized using gradient descent to maximize the likelihood of the training data.

\textbf{A linear graph-based parser (LGPar)} uses perceptron-like online training to directly learn $\textbf{w}$. The workflow is similar to Algorithm \ref{alg:training-constrained}, except that the gold-standard reference $\mathbf{d^+}$ is directly provided in the training data without the need of constrained decoding in line \ref{alg:ln:constrained-gpar}. 
Previous work mostly adopts linear models to build dependency parsers since perceptron training is simple yet effective in achieving competitive parsing accuracy in variety of languages. Recently, LLGPar attracts more attention due to its capability in producing subtree probabilities and learning from PA \cite{zhenghua-c14,xuezhe-ma-15-crf-parsing}.

\subsection{Transition-based Approach}\label{sec:transition-dp}

The transition-based method builds a dependency by applying sequence of shift/reduce actions $\mathbf{a}$, and factorizes the score of a tree into the sum of scores of each action in $\mathbf{a}$ \cite{YamadaMatsumoto2003,Nivre-iwpt03,zhang-nivre-p11-transition}: 
\begin{equation}\label{eq:trans-dp}
\begin{split}
Score(\mathbf{x},\mathbf{d}; \mathbf{w}) & = Score 
{(\mathbf{x}, \mathbf{a} \rightarrow \mathbf{d}; \mathbf{w})} \\
& = \sum\nolimits
_{i=1}^{|\mathbf{a}|}{Score(\mathbf{x},c_{i},a_{i};\mathbf{w})}
\end{split}
\end{equation}
where $\mathbf{a}_i$ is the action taken at step $i$ and ${c}_i$ is the configuration status 
after taking action ${a}_1...{a}_{i-1}$.
State-of-the-art transition-based methods usually use inexact beam search to 
find a highest-scoring action sequence, and adopt global perceptron-like training to learn $\mathbf{w}$.
We build an arc-eager transition-based dependency parser and the state-of-the-art features described in \cite{zhang-nivre-p11-transition}, referred as
\textbf{a linear transition-based parser (LTPar)}.

\section{Directly training parsers with PA} \label{sec:train-partial}

As described in \newcite{zhenghua-c14}, \textbf{LLGPar} can naturally learn from PA based on the idea of ambiguous labeling, which allows a sentence to have multiple parse trees (forest) as its gold-standard reference \cite{riezler-02-ambiguous-labelings,dredze-09-sequence-learning-multiple-labels,tackstrom-mcdonald-nivre:2013:NAACL-HLT}.
First, a partial tree $\mathbf{d}^p$ is converted in to a forest by adding all possible dependencies pointing to remaining words without heads, with the constraint that a newly added dependency does not violate existing ones in $\mathbf{d}^p$.
The forest can be formally defined as $\mathcal{F}(\mathbf{x}, \mathbf{d}^p)=\{\mathbf{d}: \mathbf{d} \in \mathcal{Y}(\mathbf{x}), 
 \mathbf{d}^p \subseteq \mathbf{d}\}$, whose conditional probability is the sum of probabilities of all trees that it contains:
\begin{equation} \label{eq:forest-prob}
p(\mathbf{d}^p|\mathbf{x};\mathbf{w}) = \sum_{\mathbf{d} \in \mathcal{F}(\mathbf{x}, \mathbf{d}^p)} p(\mathbf{d}|\mathbf{x};\mathbf{w})
\end{equation}
Then, we can define an forest-based training objective function to maximize the likelihood of training data as described in \newcite{zhenghua-c14}.

\begin{algorithm}[tb]
\centering
\caption{Perceptron training based on constrained decoding.}\label{alg:training-constrained}
\begin{small}
\begin{algorithmic}[1]
\STATE \textbf{Input:} Partially labeled data $\mathcal{D}=\{(\mathbf{x}_j,\mathbf{d}^p_j)\}_{j=1}^{N}$; 
\textbf{Output:} $\mathbf{w}$;
\textbf{Initialization: $\mathbf{w}^{(0)}=\mathbf{0}$, $k=0$}
\FOR [\textsl{iterations}]{$i=1$ \TO $I$}
	\FOR [\textsl{traverse}] {$(\mathbf{x}_j,\mathbf{d}^p_j) \in \mathcal{D}$}
	
		\STATE 
		$\mathbf{d^-} = \argmax\nolimits_{\mathbf{d} \in \mathcal{Y}(\mathbf{x}_j)} {Score(\mathbf{x}_j, \mathbf{d};\mathbf{w})}$ \COMMENT{\textsl{Unconstrained decoding: LGPar}} \label{alg:ln:unconstrained-gpar}
		\STATE $\mathbf{a^-} = \argmax\nolimits_{\mathbf{a} \rightarrow \mathbf{d} \in \mathcal{Y}(\mathbf{x}_j)} {Score(\mathbf{x}_j, \mathbf{a} \rightarrow \mathbf{d}; \mathbf{w})}$ \COMMENT{\textsl{Unconstrained decoding: LTPar}} \label{alg:ln:unconstrained-tpar}
		\STATE 
		$\mathbf{d^+} = \argmax\nolimits_{\mathbf{d} \in \mathcal{Y}(\mathbf{x}_j, \mathbf{d}^p_j)} {Score(\mathbf{x}_j, \mathbf{d};\mathbf{w})}$ \COMMENT{\textsl{Constrained decoding: LGPar}} \label{alg:ln:constrained-gpar}
		\STATE $\mathbf{a^+} = \argmax\nolimits_{\mathbf{a} \rightarrow \mathbf{d} \in \mathcal{Y}(\mathbf{x}_j, \mathbf{d}_j^p)} {Score(\mathbf{x}_j, \mathbf{a} \rightarrow \mathbf{d}; \mathbf{w})}$ \COMMENT{\textsl{Constrained decoding: LTPar}} \label{alg:ln:constrained-tpar}
		\STATE $\mathbf{w}_{k+1} = \mathbf{w}_{k} + \mathbf{f}(\mathbf{x},\mathbf{d^+}) - \mathbf{f}(\mathbf{x},\mathbf{d^-})$ \COMMENT{\textsl{Update: LGPar}} \label{alg:ln:update-gpar}	
		\STATE $\mathbf{w}_{k+1} = \mathbf{w}_{k} + \mathbf{f}(\mathbf{x},\mathbf{a^+}) - \mathbf{f}(\mathbf{x},\mathbf{a^-})$ \COMMENT{\textsl{Update: LTPar}}	\label{alg:ln:update-tpar}	
		\STATE $k=k+1$
	\ENDFOR
\ENDFOR
\end{algorithmic}
\end{small}
\end{algorithm}

\textbf{LGPar} can be extended to directly learn from PA based on the idea of constrained decoding, as shown in Algorithm \ref{alg:training-constrained}, 
which has been previously applied to Chinese word segmentation with partially labeled sequences \cite{jiang-wenbin-p10-bilingual-projection}.
The idea is using the best tree $\mathbf{d^+}$ in the constrained search space $\mathcal{Y}(\mathbf{x}_j, \mathbf{d}^p_j)$ (line \ref{alg:ln:constrained-gpar}) as a pseudo gold-standard reference for weight update.
In traditional perceptron training, $\mathbf{d^+}$ would be a complete parse tree provided in the training data.
It is trivial to implement constrained decoding for graph-based parsers, and we only need to  
disable some illegal combination operations during dynamic programming.

\textbf{LTPar} can also directly learn from PA in a similar way, as shown in Algorithm \ref{alg:training-constrained}.
Constrained decoding is performed to find a pseudo gold-standard reference (line \ref{alg:ln:constrained-tpar}).
It is more complicate to design constrained decoding for transition-based parsing than graph-based parsing.
Fortunately, \newcite{nivre-j14-constrained} propose a procedure to enable arc-eager parsers
to decode in the search space constrained by some given dependencies. We ignore the details due to the space limitation.


\section{Experiments}

\subsection{Data, parameter settings, and evaluation metric}

\setlength{\tabcolsep}{5pt}
\begin{table}[tb]
\begin{small}
\begin{center}
\begin{tabular}{ c |  *{2}{r} | r | r   }
& train-$1K$ 	& train-$39K$ & development & test \\
\hline
Sentence Number & 1,000 & 38,832 & 1,700 & 2,416 \\
Token Number & 24,358 & 925,670 & 40,117 & 56,684 \\
\hline
\end{tabular}
\end{center}
\end{small}
\caption{Data Statistics. FA is always used for train-$1K$, whereas PA is simulated for train-$39K$.}
\label{tbl:data-stat}
\end{table}

We conduct experiments on Penn Treebank (PTB), and 
follow the standard data split data (sec $2$-$21$ as training, sec $22$ as development, and sec $23$ as test). 
Original bracketed structures are converted into dependency structures using Penn2Malt with default head-finding rules.
We build a CRF-based bigram part-of-speech (POS) tagger to produce automatic POS tags for all train/dev/test data (10-way jackknifing on training data), with tagging accuracy $97.3\%$ on test data.
As suggested by an earlier anonymous reviewer, we further split the training data into two parts.
We assume that the first $1K$ training sentences are provided as a small-scale data with FA, which can be obtained by a small amount of manual annotation or through cross-lingual projection methods.
We simulate PA for the remaining $39K$ sentences. 
Table \ref{tbl:data-stat} shows the data statistics.

We train LLGPar with stochastic gradient descent \cite{finkel-kleeman-manning:2008:ACLMain}.
We set the beam size to $64$ during both training and evaluation of LTPar.
Following standard practice established by \newcite{collins-emnlp02-perc}, we adopt averaged weights for evaluation of LGPar and LTPar and use the early-update strategy during training LTPar.

Since we have two sets of training data, we adopt the simple corpus-weighting strategy of \newcite{zhenghua-c14}. In each iteration, we merge train-$1K$ and a subset of random $10K$ sentences from train-$39K$, shuffle them, and then use them for training. 
For all parsers, training terminates when the peak parsing accuracy on dev data does not improve in $30$ consecutive iterations.
For evaluation, we use the standard unlabeled attachment score (UAS) excluding punctuation marks.

\subsection{Three settings for simulating PA on train-$39K$ }

In order to simulating PA for each sentence in train-$39K$, we only keep $\alpha\%$ gold-standard dependencies (not considering punctuation marks), and remove all other dependencies.
We experiment with three simulation settings to fully investigate the capability of different approaches in learning from PA.

\textbf{Random} ($30\%$ or $15\%$): For each sentence in train-$39K$, we randomly select $\alpha\%$ words, and only keep dependencies linking to these words. 
With this setting, we aim to purely study the issue without biasing to certain structures. 
This setting may be best fit the scenario automatic syntax projection based on bitext, where the projected dependencies tend to be arbitrary (and noisy) due to the errors in automatic source-language parses and word alignments and non-isomorphism syntax between languages. 

\textbf{Uncertain} ($30\%$ or $15\%$): In their work of active learning with PA, \newcite{zhenghua-p16} show that the marginal probabilities from LLGPar is the most effective uncertainty measurement for selecting the most informative words to be annotated. 
Following their work, we first train LLGPar on train-$1K$ with FA, and then use LLGPar to parse train-$39K$ and select $\alpha\%$ most uncertain words to keep their heads.

Following \newcite{zhenghua-p16}, we measure the uncertainty of a word $w_i$ according to the \emph{marginal probability gap} between its two most likely heads $h_i^0$ and $h_i^1$.   
\begin{equation} \label{eq:partial-marg-prob-gap}
\textit{Uncertainty}(\mathbf{x}, i) = p(h_i^0 \curvearrowright i|\mathbf{x})-p(h_i^1 \curvearrowright i|\mathbf{x})
\end{equation}
The intuition is that the smaller the probability gap is, the more uncertain the model is about $w_i$. 
The marginal probability of a dependency is the sum of probabilities of all 
legal trees that contain the dependency.
\begin{equation} \label{eq:marg-prob}
p(h \curvearrowright m|\mathbf{x}) = \sum_{
\substack{\mathbf{d} \in \mathcal{Y}(\mathbf{x})}: h \curvearrowright m \in \mathbf{d}
 } p(\mathbf{d}|\mathbf{x})
\end{equation}


This setting fits the scenario of active learning, which aims to save annotation effort by only annotating the most useful structures.
From another perspective, this settings may tend to bias to LLGPar by keeping structures that are most useful for LLGPar.

\textbf{Divergence} ($13.68\%$): We train all three parsers on train-$1K$, and use them to parse train-$39K$. If their output trees do not assign the same head to a word, then we keep the gold-standard dependency pointing to the word, leading to $13.68\%$ remaining dependencies.
Different from the uncertain setting, this setting does not bias to any parser.

\subsection{Results of different parsers trained on FA}

\setlength{\tabcolsep}{5pt}
\begin{table}[tb]
\begin{small}
\begin{center}
\begin{tabular}{ c |  *{7}{c}   }
& LLGPar	& LGPar & LTPar & BerkeleyParser & TurboParser & Mate-tool & ZPar \\
\hline
on Dev & \textbf{93.16} & 93.00 & 92.77 & 92.84 & 92.86 & 92.58 & 92.42 \\
on Test & 92.42 & 92.43 & 92.01 & \textbf{92.85} & 92.63 & 92.48 & 92.12 \\
\hline
\end{tabular}
\end{center}
\end{small}
\caption{UAS of different parsers trained on all training data ($40K$)}.
\label{tbl:res-diff-parser-on-FA}
\end{table}

We train the three parsers on all the training data with FA. 
We also employ four publicly available parsers with their default settings. 
BerkeleyParser (v1.7) is a constituent-structure parser, whose results are converted into dependency structures \cite{petrov-klein-naacl07}.
TurboParser (v2.1.0) is a linear graph-based dependency parser using linear programming for inference \cite{martins-p13-turboparser}. 
Mate-tool (v3.3) is a linear graph-based dependency parser very similar to our implemented LGPar \cite{bohnet-C10-hash-kernel}.
ZPar (v0.6) is a linear transition-based dependency parser very similar to our implemented LGPar \cite{ZhangYue-J11-perceptron-beam-search}.
The results are shown in Table \ref{tbl:res-diff-parser-on-FA}.
We can see that the three parsers that we implement achieve competitive parsing accuracy and serve as strong baselines.

\subsection{Results of the directly-train approaches}

\setlength{\tabcolsep}{2pt}
\begin{table}[tb]
\begin{small}
\begin{center}
\begin{tabular}{ c | *{3}{l} | *{2}{l} | *{2}{l} | *{1}{l}  }
& \multicolumn{3}{|c}{FA (random)} & \multicolumn{2}{|c}{PA (random)} & \multicolumn{2}{|c}{PA (uncertain)} & \multicolumn{1}{|c}{PA (divergence)} \\ 
\cline{2-9} 
& $100\%$  & $30\%$ & $15\%$  & $30\%$ & $15\%$ & $30\%$ & $15\%$ & $13.68\%$ \\
\hline
LLGPar	
& \textbf{93.16}
& \textbf{91.93}	& \textbf{91.15}
& \textbf{92.39}	& \textbf{91.66}
& \textbf{93.02} & \textbf{92.44}
& \textbf{92.42} \\
LGPar
& 93.00 (-0.16)
& 91.76 (-0.17) & 90.80 (-0.35)
& 91.63 (-0.76) & 90.62	(-1.04)
& 92.46 (-0.56) & 91.64	(-0.80)
& 91.69 (-0.73) \\
LTPar	
& 92.77 (-0.39)
& 91.22 (-0.71) & 90.35	(-0.80)
& 91.12	(-1.27) & 90.12	(-1.54)
& 91.35 (-1.67) & 90.99	(-1.45) 
& 90.70 (-1.72) \\
\hline
\end{tabular}
\end{center}
\end{small}
\caption{UAS on dev data: parsers are directly trained on train-$1K$ with FA and train-$39K$ with PA. 
``FA (random) $\alpha\%$'' means randomly selecting $\alpha\%$ sentences with FA from train-$39K$ for training. Numbers in parenthesis are the accuracy gap below the corresponding LLGPar.
} 
\label{tbl:res-directly-trained}
\end{table}

The three parsers are directly trained on train-$1K$ with FA and train-$39K$ with PA based on the methods described in Section \ref{sec:train-partial}.
Table \ref{tbl:res-directly-trained} shows the results.

\textbf{Comparing the three parsers}, we have several clear findings.
(1) LLGPar achieves best performance over all settings and is very effective in learning from PA.
(2) The accuracy gap between LGPar and LLGPar becomes larger with PA than with FA, indicating LGPar is less effective in learning from PA than LLGPar.
(3) LTPar lags behind LLGPar by large margin and is ineffective in learning from PA.

\textbf{FA (random) vs. PA (random)}: from the results in the two major columns, we can see that LLGPar achieves higher accuracy by about $0.5\%$ when trained on sentences with $\alpha\%$ random dependencies than when trained on $\alpha\%$ random sentences with FA. 
This is reasonable and can be explained under the assumption that LLGPar can make full use of PA in model training.
In fact, in both cases, the training data contains approximately the same number of annotated dependencies. However, from the perspective of model training, given some dependencies in the case of PA, more information about the syntactic structure can be derived.\footnote{Also, as suggested in the work of \newcite{zhenghua-p16}, annotating PA is more time-consuming than annotating FA in terms of averaged time for each dependency, 
since dependencies in the same sentence are correlated and earlier annotated dependencies usually make later annotation easier.} 
Taking Figure \ref{fig:example-partial-tree} as an example, ``I$_1$'' can only modify ``saw$_2$'' due to the single-root and single-head constraints;
similarly, ``Sarah$_3$'' can only modify either ``saw$_2$'' or ``with$_2$''; and so on. 
Therefore, given the same amount of annotated dependencies, random PA contains more syntactic information than random FA, which explains why LLGPar performs better with PA than FA.

In contrast, both LGPar and LTPar achieve slight lower accuracy with PA than with FA. 
This is another evidence that LGPar and LTPar is less effective than LLGPar in learning from PA. 

\textbf{PA (random) vs. PA (uncertain)}: we can see that all three parser achieves much higher accuracy in the latter case.\footnote{The only exception is LTPar with $30\%$ PA, the accuracy increases by only $91.35-91.12=0.23\%$, which may be caused by the ineffectiveness of LTPar in learning from PA. 
} 
The annotated dependencies in PA (uncertain) are most uncertain ones for current statistical parser (i.e., LLGPar), and thus are more helpful for training the models than those in PA (random).
Another phenomenon is that, in the case of PA (uncertain), increasing $\alpha\%=15\%$ to $30\%$ actually doubles the number of annotated dependencies, but only boost accuracy of LLGPar by $93.02-92.44=0.58\%$, which indicates that newly added $15\%$ dependencies are much less useful since the model can already well handle these low-uncertainty dependencies.

\textbf{PA (uncertain, $15\%$) vs. PA (divergence)}: we can see that the all three parsers achieve similar parsing accuracies. This indicates that uncertainty measurement based on LLGPar can actually discovers useful dependencies to be annotated without particularly biasing towards itself.

In summary, we can conclude from the results that \emph{LLGPar can effectively learn from PA, whereas LGPar is slightly less effective and LTPar is ineffective at all.}

\subsection{Results of the complete-then-train methods}

\setlength{\tabcolsep}{3pt}
\begin{table}[tb]
\begin{small}
\begin{center}
\begin{tabular}{ c | *{1}{c} | *{2}{l} | *{2}{l} | *{1}{l}  }
\multirow{2}{*}{Parser for completion}
& \multicolumn{1}{|c}{No constraints} & \multicolumn{2}{|c}{PA (random)} & \multicolumn{2}{|c}{PA (uncertain)} & \multicolumn{1}{|c}{PA (divergence)} \\
\cline{2-7}
 & $0\%$ & $30\%$ & $15\%$ & $30\%$ & $15\%$ & $13.68\%$ \\ 
\cline{1-7} 
LLGPar-$1K$ &
\textbf{86.67} &
\textbf{92.65} (+5.98) & \textbf{90.02} (+3.35) &
\textbf{97.43} (+10.76) & \textbf{94.43} (+7.76) &
\textbf{94.36} (+7.69) \\
LGPar-$1K$ & 
86.05&
92.16 (+6.11) & 89.48 (+3.43) &
97.30 (+11.25) & 94.11 (\textbf{+8.06}) &
94.21 (+8.16) \\
LTPar-$1K$ &
85.38&
91.76 (\textbf{+6.38}) & 88.89 (\textbf{+3.51}) & 
96.90 (\textbf{+11.52}) & 93.35 (+7.97) & 
93.85 (\textbf{+8.47}) \\
\hline
LLGPar-$1K$+$39K$ &
-- &
\textbf{95.55} (+2.90) &	\textbf{93.37} (+3.35) & 
\textbf{98.30} (+0.87) &	\textbf{96.22} (+1.79) &	
\textbf{95.57} (+1.21) \\
\hline
\end{tabular}
\end{center}
\end{small}
\caption{UAS of full trees in train-$39K$ completed via constrained decoding.
} 
\label{tbl:res-constrained-decoding}
\end{table}

The most straight-forward method for learning from PA is the complete-then-learn method \cite{mirroshandel-iwpt-2011-partial-annotation}. 
The idea is first using an existing parser to complete partial trees in train-$39K$ into full trees based on constrained decoding, and then training the target parser on train-$1K$ with FA and train-$39K$ with completed FA.

\textbf{Results of completing via constrained decoding: }
Table \ref{tbl:res-constrained-decoding} reports UAS of the completed trees on train-$39K$ using two different strategies for completion. ``No constraints ($0\%$)'' means that train-$39K$ has no annotated dependencies and normal decoding without constraints is used. In the remaining columns, each parser performs constrained decoding on PA where $\alpha\%$ dependencies are provided in each sentence.
\begin{itemize}
\item \textbf{Coarsely-trained-self for completion}: We complete PA into FA using corresponding parsers coarsely trained on only train-$1K$ with FA. We call these parsers \emph{LLGPar-$1K$, LLTPar-$1K$, LTPar-$1K$} respectively. 
\item \textbf{Fine-trained-LLGPar for completion}: We complete PA into FA using LLGPar fine trained on both train-$1K$ with FA and train-$39K$ with PA. We call this LLGPar as \emph{LLGPar-$1K$+$39K$}. Please note that \emph{LLGPar-$1K$+$39K$} actually performs \emph{closed test} in this setting, meaning that it parses its training data. For example, 
\emph{LLGPar-$1K$+$39K$} trained on random ($30\%$) is employed to complete the same data by filling the remaining $70\%$ dependencies. 
\end{itemize}
Comparing the three parsers trained on train-$1K$, we can see that constrained decoding has similar effects on all three parsers, and is able to return much more accurate trees. Numbers in parenthesis show the accuracy gap between normal ($0\%$) and constrained decoding. 
This suggests that constrained decoding itself is not responsible for the ineffectiveness of Algorithm \ref{alg:training-constrained} for LTPar.

Comparing the results of LLGPar-$1K$ and LLGPar-$1K$+$39K$ (numbers in parenthesis showing the accuracy gap), it is obvious that the latter produces much better full trees since the fine-trained LLGPar can  make extra use of PA in train-$39K$ during training. 

\textbf{Results of training on completed FA: } Table \ref{tbl:res-completed-FA} compares performance of the three parsers trained on train-$1K$ with FA and train-$39K$ with completed FA, from which we can 
draw several clear and interesting findings.
First, different from the case of directly training on PA, the three parsers achieve very similar parsing accuracies when trained on data with completed FA in both completion settings. 
Second, \emph{using parsers coarsely-trained on train-$1K$ for completion leads to very bad performance}, which is even much worse than those of the directly-train method in Table \ref{tbl:res-directly-trained} except for LTPar with uncertain ($30\%$).
Third, \emph{using the fine-trained LLGPar-$1K$+$39K$ for completion makes LGPar and LTPar achieve nearly the same accuracies with LLGPar}, which may be because LLGPar provides complementary effects during completion, analogous to the scenario of co-training.

\setlength{\tabcolsep}{5pt}
\begin{table}[tb]
\begin{small}
\begin{center}
\begin{tabular}{c | *{2}{c}  | *{2}{c} | *{1}{c} || *{2}{c}  | *{2}{c} | *{1}{c} }
\multicolumn{1}{c}{} 
& \multicolumn{5}{|c||}{Completed by LLGPar/LGPar/LTPar-$1K$  } 
& \multicolumn{5}{c}{Completed by LLGPar-$1K$+$39K$  } \\
\cline{2-11}
\multicolumn{1}{c}{} 
& \multicolumn{2}{|c}{PA (random)} & \multicolumn{2}{|c}{PA (uncertain)} & \multicolumn{1}{|c||}{PA (divergence)} 
& \multicolumn{2}{c}{PA (random)} & \multicolumn{2}{|c}{PA (uncertain)} & \multicolumn{1}{|c}{PA (divergence)}\\
\cline{2-11}
& $30\%$ & $15\%$ & $30\%$ & $15\%$ & $13.68\%$ & $30\%$ & $15\%$ & $30\%$ & $15\%$ & $13.68\%$\\ 
\cline{1-11} 
LLGPar &
\textbf{89.91} &	\textbf{88.69} &	
\textbf{92.05} &	\textbf{90.77} &	
\textbf{91.05} & 
\textbf{92.29} &	91.54& 
\textbf{92.86} &	\textbf{92.33} &
\textbf{92.33}
\\
LGPar & 
89.42&	88.32&	
91.85&	90.66&	
90.68&
92.17&	\textbf{91.59} &
92.84&	92.21&	
92.19
\\
LTPar &
89.17&	87.72&	
91.59&	90.12&	
90.54&
92.05&	91.37&	
92.42&	92.10&	
92.01
\\
\hline
\end{tabular}
\end{center}
\end{small}
\caption{UAS on dev data: parsers are trained on train-$1K$ with FA and train-$39K$ with completed FA.
} 
\label{tbl:res-completed-FA}
\end{table}

\setlength{\tabcolsep}{5pt}
\begin{table}[tb]
\begin{small}
\begin{center}
\begin{tabular}{c | *{2}{c}  | *{2}{c} | *{1}{c} || *{2}{c}  | *{2}{c} | *{1}{c} }
\multicolumn{1}{c}{} 
& \multicolumn{5}{|c||}{Directly train on train-$39K$ with PA } 
& \multicolumn{5}{c}{Train-$39K$ with FA completed by LLGPar-$1K$+$39K$ } \\
\cline{2-11}
\multicolumn{1}{c}{} 
& \multicolumn{2}{|c}{PA (random)} & \multicolumn{2}{|c}{PA (uncertain)} & \multicolumn{1}{|c||}{PA (divergence)} 
& \multicolumn{2}{c}{PA (random)} & \multicolumn{2}{|c}{PA (uncertain)} & \multicolumn{1}{|c}{PA (divergence)}\\
\cline{2-11}
& $30\%$ & $15\%$ & $30\%$ & $15\%$ & $13.68\%$ & $30\%$ & $15\%$ & $30\%$ & $15\%$ & $13.68\%$\\ 
\cline{1-11} 
LLGPar &
\textbf{91.73}&	\textbf{91.02}&	\textbf{92.34}&	\textbf{91.83}&	\textbf{91.84}&
91.46&	\textbf{90.99}&	\textbf{92.20}&	\textbf{91.59}&	91.53
\\
LGPar & 
91.17&	90.36&	91.99&	91.28&	91.22&
\textbf{91.55}&	90.96&	91.98&	91.57&	\textbf{91.56}
\\
LTPar &
90.79&	89.89&	90.47&	90.37&	90.06&
91.48&	90.78&	91.80&	91.45&	91.52
\\
\hline
\end{tabular}
\end{center}
\end{small}
\caption{UAS on test data: comparison of the directly-train and complete-then-train methods.
} 
\label{tbl:res-test-comparison}
\end{table}

\subsection{Results on test data: directly-train vs. complete-then-train}

Table \ref{tbl:res-test-comparison} reports UAS on test data of parsers directly trained on train-$1K$ with FA and train-$39K$ with PA, and of those trained on train-$1K$ with FA and train-$39K$ with FA completed by fine-trained LLGPar-$1K$+$39K$.
The results are consistent with the those on dev data in Table \ref{tbl:res-directly-trained} and \ref{tbl:res-completed-FA}.
Comparing the two settings, we can draw two interesting findings. 
First, LLGPar performs slightly better with the directly-train method.
Second, LGPar performs slightly better with the complete-then-train method in most cases except for uncertain ($30\%$).
Third, LTPar performs much better with the complete-then-train method.

\section{Failed attempts to enhancing LTPar}

All experimental results in the previous section suggest that LTPar is ineffective in learning from PA, and Table \ref{tbl:res-constrained-decoding} indicates that constrained decoding itself works well for LTPar. 
In contrast, LGPar is also based on constrained decoding and works much better than LTPar. 
The most important difference is that in line \ref{alg:ln:unconstrained-gpar}-\ref{alg:ln:constrained-tpar} of Algorithm \ref{alg:training-constrained}, LGPar uses dynamic programming based exact search algorithm to find the highest-scoring tree according to the current model, whereas LTPar use approximate beam search algorithm. The approximate search procedure may cause the optimal tree drops off the beam too soon and thus the returned $\mathbf{a^+}$ may cause the model be updated to bias to certain wrong structures, which cannot be further covered due to the lack of sufficient supervision in the scenario of PA. 

We have tried three strategies to enhance LTPar so far though little progress has been made. 
First, we set the beam size to $32/64/128/256$, and hope LTPar may learn better from PA with larger beam. 
Second, as suggested by an earlier anonymous reviewer, we use $k$-best $\mathbf{a^+}$ and $\mathbf{a^-}$ instead of the 1-best outputs for feature weight update. 
We try to use the averaged feature vector of $k$-best $\mathbf{a^+}$ and/or $k$-best $\mathbf{a^-}$ in line \ref{alg:ln:update-tpar}.
Third, we also try a conservative update strategy. The idea is that first we obtain $\mathbf{a^-}$ (corresponding to a tree $\mathbf{d^-}$) in line \ref{alg:ln:unconstrained-tpar}. Then, for each dependency in $\mathbf{d^-}$ that is compatible with those in the partial tree $\mathbf{d}_j^p$, we temporarily insert it into $\mathbf{d}_j^p$. We use the enlarged $\mathbf{d}_j^p$ in line \ref{alg:ln:constrained-tpar}. 
In this way, the returned $\mathbf{d^+}$ is more similar to $\mathbf{d^-}$ so that less risk is taken during model update. 
So far, the results for all three strategies are negative. However, we will keep looking into this issue in future. We will give more detailed descriptions and results upon publication with extra space.

\section{Related work} \label{sec:related-work}

In parsing community, most previous works adopt ad-hoc methods to learn from PA. 
\newcite{sassano-p10-partial-annotation}, \newcite{jiang-wenbin-p10-bilingual-projection}, and \newcite{flannery-iwpt15-active-learning-partial-annotation} 
convert partially annotated instances into local dependency/non-dependency classification instances, which may suffer from the lack of non-local correlation between dependencies in a tree.

\newcite{mirroshandel-iwpt-2011-partial-annotation} and \newcite{majidi-13-active-learning-committee} adopt the complete-then-learn method. They use parsers coarsely trained on existing data with FA for completion via constrained decoding. 
However, our experiments show that this leads to dramatic decrease in parsing accuracy. 

\newcite{nivre-j14-constrained} present a constrained decoding procedure for arc-eager transition-based parsers. However, their work focuses on allowing their parser to effectively exploit external constraints during the evaluation phase. In this work, we directly employ their method and show that constrained decoding is effective for LTPar and thus irresponsible for its ineffectiveness in learning PA.

Directly learning from PA based on constrained decoding is previously proposed by \newcite{jiang-p13-natural-annotation} for Chinese word segmentation, which is treated as a character-level sequence labeling problem. In this work, we first apply the idea to LGPar and LTPar for directly learning from PA.

Directly learning from PA based on a forest-based objective in LLGPar is first proposed by \newcite{zhenghua-c14}, inspired by the idea of \emph{ambiguous labeling}. Similar ideas have been extensively explored recently in sequence labeling tasks \cite{liu-d14-crf-natural-annotation,yang-vozila-d14-crf-partial-annotation,marcheggiani-d14-active-learning-partial-annotation}.

\newcite{hwa-99-partial-annotation} pioneers the idea of exploring PA for constituent grammar induction based on a variant Inside-Outside re-estimation algorithm \cite{pereira-92-inside-outside}.
\newcite{clark-curran-06-partial-annotation} propose to train a Combinatorial Categorial Grammar parser using partially labeled data only containing predicate-argument dependencies. 
\newcite{mielens-sun-baldridge:2015:ACL:parser-imputation} propose to impute missing dependencies based on Gibbs sampling in order to enable traditional parsers to learn from partial trees.

\section{Conclusions}\label{sec:con}

This paper investigates the problem of training dependency parsers on partially labeled data. Particularly, we focus on the realistic scenario where we have a small-scale training dataset with FA and a large-scale training dataset with PA.
We experiment with three settings for simulating PA.
We compare several directly-train and complete-then-train approaches with three mainstream parsers, i.e., 
LLGPar, LGPar, and LTPar.
Finally, we draw the following important conclusions.
(1) For the complete-then-train approach, using parsers coarsely trained on small-scale data with FA for completion leads to unsatisfactory results.
(2) LLGPar is able to make full use of PA for training.
In contrast, LGPar is slightly inferior  and LTPar performs badly in learning from PA.
(3) The complete-then-train approach can make LGPar and LTPar on par with LLGPar in terms of parsing accuracy if using LLGPar fine trained on all data with both FA and PA for completion.

For future, we will further investigate the reason behind the ineffectiveness of LTPar in learning from PA, and try to propose effective strategies to solve the issue. Our next plan is to employ the dynamic programming-enhanced beam search by merging equivalent states proposed by \newcite{huang-sagae:2010:ACL}, which allows the parser to explore larger search space during decoding.
Moreover, we also plan to consider more constraints beyond dependencies. For example, \newcite{nivre-j14-constrained} propose a constrained decoding procedure which can also incorporate bracketing constraints, i.e., a certain span forming a single-root subtree, which would be interesting yet challenging for graph-based parsers due to the complexity of designing dynamic programming based algorithms.






\section*{Acknowledgments}
The authors would like to thank the anonymous reviewers for the helpful comments.

\bibliographystyle{acl}

\bibliography{../ref/reference}

\end{document}